\documentclass[twoside]{article}
\usepackage{aistats2018}
\usepackage{amsmath}
\usepackage{amsfonts}
\usepackage{graphicx}
\usepackage{amsthm}

\begin{document}

%

%

\twocolumn[

\aistatstitle{Compressing Deep Neural Networks: A New Hashing Pipeline Using Kac's Random Walk Matrices}

\aistatsauthor{Jack Parker-Holder \And Sam Gass}

\aistatsaddress{Columbia University \And  Columbia University} ]

\begin{abstract}
The popularity of deep learning is increasing by the day. However, despite the recent advancements in hardware, deep neural networks remain computationally intensive. Recent work has shown that by preserving the angular distance between vectors, random feature maps are able to reduce dimensionality without introducing bias to the estimator. We test a variety of established hashing pipelines as well as a new approach using Kac's random walk matrices.  We demonstrate that this method achieves similar accuracy to existing pipelines.

\end{abstract}

\section{Introduction}

After decades of research, deep learning methods burst onto the scene with the success of AlexNet in 2012 \cite{Alexnet}, the first convolutional neural network to win ImageNet Large Scale Visual Classification \cite{imagenet}. Since then, there have been a series of breakthroughs, such as AlphaGo beating the world's best Go player \cite{AGO}, before being beaten 100-0 by AlphaGo Zero which learned solely by playing against itself \cite{AGZ}. Artificial Intelligence has now gone mainstream, the media have caught on and deep learning is all the rage. However, training neural networks remains computationally intensive.

Over the past few years, work on using pseudo-random matrices (i.e. matrices where some entries are indeed fully random but others are derived from them and thus not independent) has become more prominent. This is of particular interest due to the relative ease of storing such matrices. Work by Yu et al. \cite{Circ1} and Yi et al. \cite{Circ2} showed that the circulant and Gaussian Toeplitz matrices could be used to do this while Choromanska, Choromanski et al. \cite{KC} went further and proved theoretically that such methods produce unbiased estimators. 

In this paper, we explore several hashing pipelines which seek to compress the data by applying a random projection followed by the non-linear sign function. We show that these methods successfully preserve the angular distance between observations sufficiently to maintain high classification accuracy. We also test a new pipeline using Kac's random walk matrix, based onthe work of Marc Kac \cite{kac1956} in the 1950s. Recent theoretical results \cite{Kac} indicate the matrix, which produces a random rotation, can be can be constructed in $ n\log(n) $ steps. 

The structure of the paper is as follows, the next section gives an overview of the existing literature, we then explain the hashing pipelines and follow this with experimental results on the MNIST \cite{MNIST} dataset of handwritten digits.

\section{Related work}

The first work on random projections was done by Dasgupta \cite{Das} in the 1990s, who successfully applied it to real datasets \cite{Das2}. More recently, such methods have been applied to deep learning architectures (for a review see Saxe. et al, \cite{Saxe}), which is the focus of our work. 

Recently, Chen et al. \cite{Chen} introduced a neural network using hashing, which they call HashedNets, and showed that such a network was able to achieve a significant reduction in model size without the loss of accuracy.

In recent years there has been increasing interest in using pseudo-random projections to compress neural networks. Studies by Yu et al. \cite{Circ1} and Yi et al. \cite{Circ2} focused on the efficacy of the circulant Gaussian matrix and found significant gains in storage and efficiency with a minimal increase in the error rate compared to a regular neural network or an unstructured projection. 

Another recent paper by Choromanska, Choromanski et al. \cite{KC} introduce two hashing pipelines (which we will explain further later in this paper) and provide theoretical guarantees regarding concentration of the estimators around their mean. They demonstrate the accuracy of the hashing pipelines using the MNIST \cite{MNIST} dataset, and find only a small decrease in accuracy with as much as an eightfold reduction in dimensionality. They tested several structured Gaussian matrices such as the circulant and Toeplitz Gaussian matrix, and showed these projections can maintain high levels of accuracy. We seek to reproduce their results for random Gaussian matrices, as well as two structured Gaussian matrices (circulant and Toeplitz).

In addition to transformations discussed in \cite{KC}, we note exciting new work from Pillai and Smith \cite{Kac} on Kac's random walk matrices, which showed that matrices constructed using a random walk as suggested by Marc Kac \cite{kac1956} can reach a steady state in $ n\log(n) $ steps, where $n$ is the dimensionality of the data. We show that this matrix can also be included in the hashing pipeline without a reduction in accuracy. 

More details on the structure of the matrices and the processing pipeline follows in the next section. 

\section{Random Feature Maps}

\subsection{The JLT Lemma}

The goal of the transformations utilized in this paper is to input high n-dimensional vectors $x^{1}....x^{N}$ from $\mathbb{R}^N$ and compress each $x^{i}$ into a lower dimensional vector while preserving the pair-wise Euclidean distance between the vectors. The mathematical theory behind these transformations is the Johnson-Lindenstrauss Lemma \cite{JLT}: 
\newline
\newline
\smallskip
\noindent \textbf{Theorem 1: Johnson Lindenstrauss Lemma:} \textit{Let $S$ be a set of $N$ points in $\mathbb{R}^d$}. \textit{Let $A \in \mathbb{R}^{k x d} \sim iid N(0,1)$}. \textit{With a probability of at least $1 - 2N^{2} e^{-(\epsilon^{2}-\epsilon^{3})k/4}:$}

\smallskip
$$(1 - \epsilon) \| x - y \|_{2} \geq \frac{1}{\sqrt{k}} \| A(x-y)\|_{2}\leq(1+\epsilon)\|x-y\|_{2}$$
\smallskip
for any $x,y \in S$
\newline\noindent 

As k $\longrightarrow O(\frac{logN}{\epsilon^{2}})$, the probability above approaches 1\cite{Circ3}.  This discovery has led to an entire field devoted to the reduction of dimensions through random projections. These transformations allow most data analyses based on Euclidean distances among points can be reduced to $O(\log n)$ times. Since its inception the technique has progressed rapidly, leading to faster machine learning algorithms in multiple fields including nearest neighbor searches and and clustering procedures.

\subsection{The hashing pipeline: unstructured matrices}

For the unstructured matrix we take a simple hashing pipeline of the matrix multiplication followed by the sign function.

$$h(x) = sign(G(x))$$

where $G$ is a Gaussian random matrix. 

\subsubsection{Gaussian random matrix}

The Gaussian random matrix is simply a matrix $G$ such that all entries $g_{i,j}$ are taken from the unit Gaussian distribution $N(0,1)$. The matrix is of the form:

$$
\begin{pmatrix}
g_{1,1} & g_{1,2} & ... &  g_{1,n} \\
g_{2,1} & g_{2,2} & ... & g_{2,n} \\
\vdots & \vdots & \vdots &  \vdots\\
g_{n,1} & g_{n,2} & ... & g_{n,n} \\
\end{pmatrix}
$$

\subsection{The hashing pipeline: structured matrices}

For structured matrices, we test two pipelines recently introduced by Choromanska, Choromanski et al \cite{KC}. They proposed two pipelines which they call extended $\Psi$-regular hashing and short $\Psi$-regular hashing, each of which consists of a pre-processing step followed by a hashing step.

\subsubsection{Extended $\Psi$-regular hashing}

The first pipeline is as follows:

$$ h(x) = sign(P.D_{2}.(H.D_{1}).X)$$

where $ D_{1} $ and $ D_{2} $ are independent copies of the diagonal matrix where each entry is taken from the set $ {-1,1} $ with probability $ \frac{1}{2} $. $H$ is the $L_{2}$-normalized Hadamard matrix. $P_{\Psi}$ is the projection matrix (in this case either $P_{circ}$ or $P_{toep}$).

\subsubsection{Short $\Psi$-regular hashing}

The second pipeline avoids applying the first random matrix $D_{1}$ as well as the Hadamard matrix:

$$ h(x) = sign(P.D.X)$$

\subsubsection{Hashing with Kac's random walk matrix}

The third pipeline we test is an adaptation of the Extended $\Psi$-regular hashing pipeline, where we replace the $HD$ block matrix with Kac's random walk matrix $M$, of the form:

$$M = B_{1}.B_{2}...B_{k}$$

where
\[
B_{i} = 
{\let\quad\thinspace
 \bordermatrix{~ & ~ & ~ & ~& i & ~ & j & ~ & ~\cr
  			~ & 1 & 0 & \cdots & ~ &  ~ & ~ & \cdots & 0  \cr
  			~ & 0 & 1 & \cdots & ~ & ~ & ~ & \cdots & 0 \cr
  			~ & \vdots  & \vdots  & \ddots & ~ & ~& ~&~&\vdots \cr
 			 i & ~ & ~ & ~& \sin\theta & ~& \cos\theta & ~& ~\cr
			 ~& ~ & ~ & ~& ~ & ~& ~ & ~& ~\cr
			 ~&~&~&~&~& \ddots&~&~& ~\cr
			  ~& ~ & ~ & ~& ~ & ~& ~ & ~&~\cr
			 j & ~ & ~ & ~& -\cos\theta & ~& \sin\theta & ~&~\cr
			 ~&~&~&~&~& ~&~&\ddots&0\cr
			 ~&0&~&~&~& ~&~&0&1 \cr}
}		 
\]

with
$$i, j \in Unif\\(0, ..., n)$$

where $n$ is the dimensionality of the dataset, and 

$$\theta \sim Unif \\( 0, 2\pi\\)$$

Recent results from Pillai and Smith \cite{Kac} showed that in order for $M$ to be a random matrix, we need to set $k$ equal to $n\log(n)$. This result makes Kac's random walk matrices an appealing component of a hash, given we seek to compress the data as efficiently as possible while preserving the angular distance between the vectors.  

Although we provide no theoretical guarantees for this pipeline, the intuition makes sense considering that each $B_{i}$ corresponds to a truly random rotation and the results from Pillai and Smith find that $n\log(n)$ steps suffice for the walk to be in a steady state. 

\subsubsection{Structured Gaussian matrices}

In our hashing pipeline we use two structured matrices, the circulant and Toeplitz Gaussian matrices. These are defined below.

\subsubsection{Circulant Gaussian matrix}

A circulant matrix $P_{circ}$ is a structured random matrix where each row of the matrix 
is a rotation of a single random vector c, where: $$c = (c_{1} ... c_{n})  \sim N(0,1)$$ 
The final result, $P_{circ}$,  is a matrix of the form given by:

$$
\begin{pmatrix}
c_{1} & c_{2} & ... &  c_{n} \\
c_{n} & c_{1} & ... & c_{n-1} \\
\vdots & \vdots & \vdots &  \vdots\\
c_{3} & c_{4} & ... & c_{2} \\
c_{2} & c_{3} & ... & c_{1} \\
\end{pmatrix}
$$

\subsubsection{Toeplitz Gaussian matrix}

A Toeplitz gaussian matrix $P_{toep}$ is a structured random matrix if each of its descending diagonals 
$ (t_{0},..,t_{n}) $ is given by: $$t_{i} \in N(0,1)$$ 
The resulting matrix is of the form:

$$
\begin{pmatrix}
t_{0} & t_{-1} & ... &  t_{-(n-1)} \\
t_{1} & t_{0} & ... & t_{-(n-2)} \\
\vdots & \vdots & \vdots &  \vdots\\
t_{n-2} & t_{n-3} & ... & t_{-1} \\
t_{n-1} & t_{n-1} & ... & t_{0} \\
\end{pmatrix}
$$

\section{Experimental results}

\subsection{The dataset}

For the purpose of our experiments, we use the MNIST dataset, which has been used extensively by researchers in the field. The dataset has 60,000 28x28 training images and 10,000 testing images, each of hand written digits between 0 and 9. Given that we are working with images we use a convolutional neural network for classification, which we will describe in the next section. 

\subsection{Neural network architecture}

The network is set up with two convolutional layers, each with filter size 3x3 and stride 1. Each convolutional layer is followed by a max pooling layer with filters of size 2x2 applied with a stride of 2. These layers are followed by fully connected layers, the first of which we fix to an arbitrary number of neurons (in this case 50), and the second of which is the output layer, with ten neurons representing the numbers 0 to 9. 

\begin{figure*}[t]
	\centering
	\includegraphics[width=150mm]{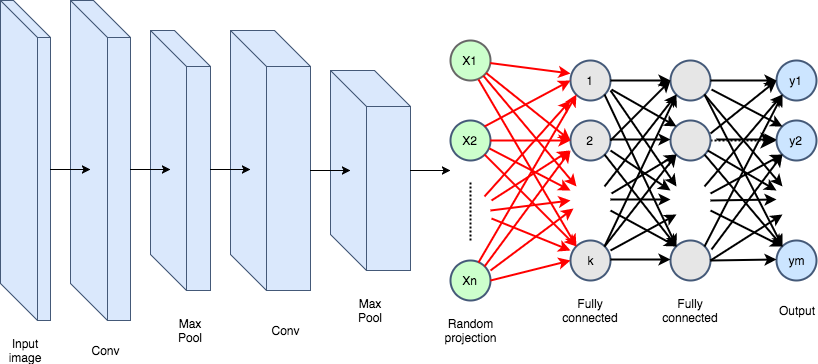}
	\caption{The neural network set-up for our experiment, where $n$ is the number of dimensions prior to the hash, $k$ is the size of the hash and $m$ is number of classes of the output (here $m=10$)}
\end{figure*}

To compress the network, we implement our hashing pipeline before the first fully connected layer, thus keeping the image in tact for the two convolutional layers while reducing dimensionality for the subsequent dense layer. In convolutional neural networks, the fully connected layers are often the bottleneck in terms of efficiency, sometimes contributing over 90\% of the storage (for example "AlexNet" \cite{paper}). 

Figure 1 shows the visual representation of the network, where the input data has $n$ dimensions (in this case $n = 1568$) and we reduce this to $k$ dimensions with the hashing step, which is represented by the red arrows. For the purpose of our experiments, tested $k = {784, 392, 196, 98, 49}$ for a dimensionality reduction ($\frac{n}{k}$) of 2, 4, 8, 16 and 32 respectively.

When training the model, we use ten epochs and the Adam optimization technique \cite{Adam}. The network was built using Keras with the TensorFlow backend. 

\subsection{Results}

\begin{figure*}[t]
	\centering
	\caption{Results for all three hashing pipelines with $P = P_{circ}$ (left) and $P = P_{toep}$ (right)}
	\includegraphics[width = 150mm]{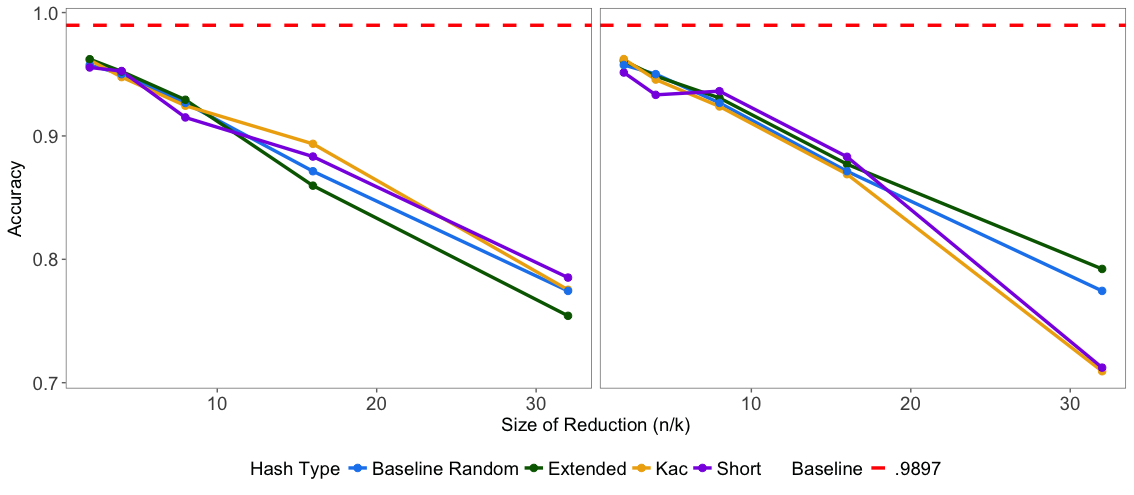}
	 \vspace{0.0cm}
\end{figure*}

We present our results in a table in two charts. For each of the three hashing pipelines mentioned in Section 3, we use both $P_{circ}$, the circulant matrix described in section 3.3.5, and $P_{toep}$, the Toeplitz Gaussian matrix described in section 3.3.6. For comparison purposes, we show the result of a "baseline" model, which is the result of training the same neural network described in Section 4.2 without a hashing step. We also present the results of the hash using the fully random Gaussian matrix, described in Section 3.2.1, which we refer to as "Random" in Table 1 and Figure 2.

In Table 1 and Figure 2 we show the size of the hash $k$, or the size of the reduction $\frac{n}{k}$ and the corresponding testing accuracy, where accuracy refers to top-1 accuracy (i.e. the percentage of the time the model predicts the correct category). As can be seen in both Figure 2 and Table 1, all three pipelines successfully preserve the angular distance between the data points to the extent that they are able to achieve high levels of accuracy. In particular, all three pipelines achieve close to 95\% accuracy for a hash size of $k = 392$, which corresponds to a 75\% compression. 

Looking at the results for the hashes using the circulant matrix, the extended pipeline using both the $HD$ block and Kac matrices performed best, achieving 96.22\% and 96.15\% accuracies respectively for $k = 784$, the smallest compression we tested. For the larger reduction in dimensionality however, the short $\Psi$-regular hashing pipeline actually did best, with over 78\% accuracy. 

For the Topelitz Gaussian matrix, all three pipelines performed well for the smallest reduction in dimensionality, achieving over 96\% accuracy. However, for the largest compression, the extended $\Psi$-regular hashing pipeline was the most robust, as the other two approaches saw large declines in accuracy to just over 70\%.

\begin{table*}[t]
\caption{Experimental results} \label{results-table}
\begin{center}
\begin{tabular}{| c | c | c c c | c c c |}
\hline
  & {\bf Random}  &  & {\bf Circulant} & &  & {\bf Toeplitz} & \\
  \hline

{\bf $k$ / $\frac{n}{k}$} & & Extended & Short & Kac & Extended & Short & Kac \\
\hline\hline
784 / 2  & 0.9575 & 0.9622 & 0.9555 & 0.9615 & 0.9609 & 0.9624 & 0.9623 \\
\hline
392 / 4  & 0.9501 & 0.9523 & 0.9526 & 0.9477 & 0.9485 & 0.9534 & 0.9457 \\
\hline
196 / 8  & 0.9268 & 0.9293 & 0.9150 & 0.9245 & 0.9308 & 0.9094 & 0.9239 \\
\hline
98 / 16  & 0.8714 & 0.8597 & 0.8833 & 0.8936 & 0.8771 & 0.8552 & 0.869 \\
\hline
49 / 32  & 0.7744 & 0.7542 & 0.7852 & 0.7753 & 0.7922 & 0.711 & 0.7093 \\
\hline
\end{tabular}
\end{center}
\end{table*}

The results we see for both the Short and Extended $\Psi$-regular hashing pipelines  are consistent with the findings in Choromanska, Choromanski et al \cite{KC}, as well as others. This work once again confirms that structured matrices, which can be stored in linear space, are almost as effective as fully random matrices in preserving the angular distance between data points. As we see here, the fully random matrix does indeed achieve strong results, however it is not obvious that the improvements warrant the increased budgeting from computational complexity that storing unstructured matrices requires.

The key result of this paper is that the pipeline which uses Kac's random walk matrix instead of the $HD$ block actually achieves comparable results, in particular for the smaller reductions in dimensionality. This is the result we hoped to see, and could prove useful for future applications given the recent theoretical results from Pillai and Smith.

\section{Conclusion}

Our results show that all three hashing pipelines are able to reduce dimensionality while preserving the angular distance between input data instances. In particular, we show that a convolutional neural network with a hashing step before the fully connected layers compares favorably with a baseline model with no hash and a hash with a fully random Gaussian matrix when classifying images on the MNIST dataset. 

We also showed Kac's random walk matrix can be used in place of the $HD$ block in the hashing pipeline to achieve an equal accuracy for significant reductions in dimensionality. The results are consistent when both the circulant and Toeplitz Gaussian matrices were used in the pipeline. 

This is an important result, as it offers potential efficiency gains which can boost the performance of practical implementations of deep neural networks, such as in robotics. Given the extent to which fully connected layers contribute to the space requirement of convolutional neural networks, it is likely that optimized versions of our proposed hashing pipeline (for example using the Fast-Fourier Transform) can drastically improve both the time and space required to train and test such networks. 

As deep learning continues to grow in popularity, approaches such as this could prove critical when using neural networks in practice.

\subsection{Acknowledgements}

We could not have done any of this work without the inspiration and guidance of Krzysztof Choromanski, who introduced us to the beauty of random feature maps. We thank him for his time and patience. 

\bibliographystyle{unsrt}

\bibliography{BDML}

\end{document}